%% file: 0-main.tex
\documentclass{article} 

\usepackage[numbers]{natbib}  

\usepackage[preprint]{neurips_2025}  
\usepackage{times}  
\usepackage{helvet}  
\usepackage{courier}  
\usepackage[hyphens]{url}  
\usepackage{graphicx} 
\usepackage{caption} 
\usepackage{algorithm}
\usepackage{algorithmic}

\usepackage{newfloat}
\usepackage{listings}
\DeclareCaptionStyle{ruled}{labelfont=normalfont,labelsep=colon,strut=off} 
\floatstyle{ruled}
\newfloat{listing}{tb}{lst}{}
\floatname{listing}{Listing}
\usepackage[utf8]{inputenc} 
\usepackage[T1]{fontenc}    
\usepackage{url}            
\usepackage{booktabs}       
\usepackage{amsfonts}       
\usepackage{nicefrac}       
\usepackage{microtype}      
\usepackage{xcolor}         
\usepackage{graphicx}

\usepackage{booktabs}
\usepackage{graphicx} 

\title{Subspace Clustering on Incomplete Data with Self-Supervised Contrastive Learning}

\author{
Huanran Li, Daniel Pimentel-Alarc\'on\\
Department of Electrical Engineering, Biostatistics\\
Wisconsin Institute of Discovery\\
University of Wisconsin-Madison\\
\texttt{\{hli488\}\{pimentelalar\}@wisc.edu}
}

\usepackage{amsmath}
\usepackage{amssymb}
\usepackage{subfigure}
\usepackage{subcaption}
\usepackage{float}

\usepackage{xcolor} 

\begin{document}

\maketitle
\begin{abstract}
Subspace clustering aims to group data points that lie in a union of low-dimensional subspaces and finds wide application in computer vision, hyperspectral imaging, and recommendation systems. However, most existing methods assume fully observed data, limiting their effectiveness in real-world scenarios with missing entries. In this paper, we propose a contrastive self-supervised framework, Contrastive Subspace Clustering (CSC), designed for clustering incomplete data. CSC generates masked views of partially observed inputs and trains a deep neural network using a SimCLR-style contrastive loss to learn invariant embeddings. These embeddings are then clustered using sparse subspace clustering. Experiments on six benchmark datasets show that CSC consistently outperforms both classical and deep learning baselines, demonstrating strong robustness to missing data and scalability to large datasets.
\end{abstract}

\section{Introduction}
\input{1-Introduction}

\section{Background}
\label{sec:background}
\input{2-Background}

\section{Model}
\label{sec:model}
\input{3-Model}

\section{Synthetic Ablation Study}
\label{sec:simulation}
\input{4-SyntheticSimulation}

\section{Experiment}
\label{sec:experiment}
\input{5-RealExperiment}

\section{Conclusion}

We proposed a novel self-supervised contrastive learning framework for subspace clustering under incomplete observations. By treating random masks as data augmentations, CSC learns invariant embeddings that preserve subspace structure without requiring reconstruction or label supervision. These embeddings can be directly used with classical clustering algorithms such as SSC, making our approach modular, scalable, and effective. Extensive experiments on six image and hyperspectral datasets demonstrate that CSC consistently outperforms classical HRMC-based techniques and modern deep learning baselines, especially at high missing rates. In future work, we plan to extend our framework to streaming and multi-view settings, and further explore theoretical guarantees for contrastive objectives under missing data.

\bibliography{ref}
\bibliographystyle{plainnat}


\end{document}

%% file: 1-Introduction.tex
Subspace clustering is a fundamental task in machine learning and signal processing that aims to cluster data points lying in a union of low-dimensional subspaces. This paradigm has found widespread applications in gene expression analysis \cite{lan2024jlonmfsc, domeniconi2004subspace, kailing2004density, zheng2020adaptive}, recommendation systems \cite{agarwal2005research, zhang2012guess, cui2021new}, image and video processing \cite{guan2024contrastive, tierney2014subspace, elhamifar2013sparse, peng2020deep, li2024grasscare}, network security anomaly detection \cite{pu2020hybrid, dromard2016online}, and financial data analytics \cite{pavlidis2006financial, li2021integrated}. However, in real-world settings, data is often incomplete due to occlusion, sensor failure, or resource constraints. Subspace clustering on incomplete data presents an important yet under-explored problem with practical significance in domains where data acquisition is costly or noisy \cite{pimentel2016group, johnson2024fusion, mao2025robust, soni2024integer, yang2015sparse}.
Despite its importance, subspace clustering with missing data is inherently challenging. Traditional subspace clustering algorithms rely heavily on complete observations to infer underlying subspace structures. When entries are missing, the data no longer lies in a clean subspace structure, and naive imputation or two-stage methods often yield suboptimal results. Moreover, classical approaches scale poorly with data size due to their reliance on pairwise affinities or matrix decompositions.

Recent advances in deep learning offer a promising avenue for overcoming the limitations of traditional subspace clustering methods in the presence of missing data \cite{ghaderi2023self, shi2023cl, fan2018matrix, monti2017geometric, li2024multi, xie2023towards}. Among these, self-supervised learning has emerged as a powerful technique for acquiring meaningful data representations without the need for explicit labels. This is particularly advantageous for subspace clustering on partially observed data, where labels are often inaccessible.
When trained with self-supervised objectives, deep networks can leverage large volumes of unlabeled and incomplete data to learn representations that are robust to both noise and missingness. These representations are capable of capturing the underlying subspace structure, even when individual feature values are absent, by generalizing across patterns observed in many partially observed instances. This stands in stark contrast to classical methods, which typically depend on pairwise similarities or complete matrix reconstruction and tend to degrade significantly under high missing rates.
Moreover, once a self-supervised model is trained, it can efficiently map new, incomplete samples to embeddings through a simple forward pass. This enables near-instantaneous inference and significantly reduces memory and computational demands compared to optimization-based methods, which often scale quadratically with the number of samples. 

\begin{figure*}
    \centering
    \includegraphics[width=0.8\linewidth]{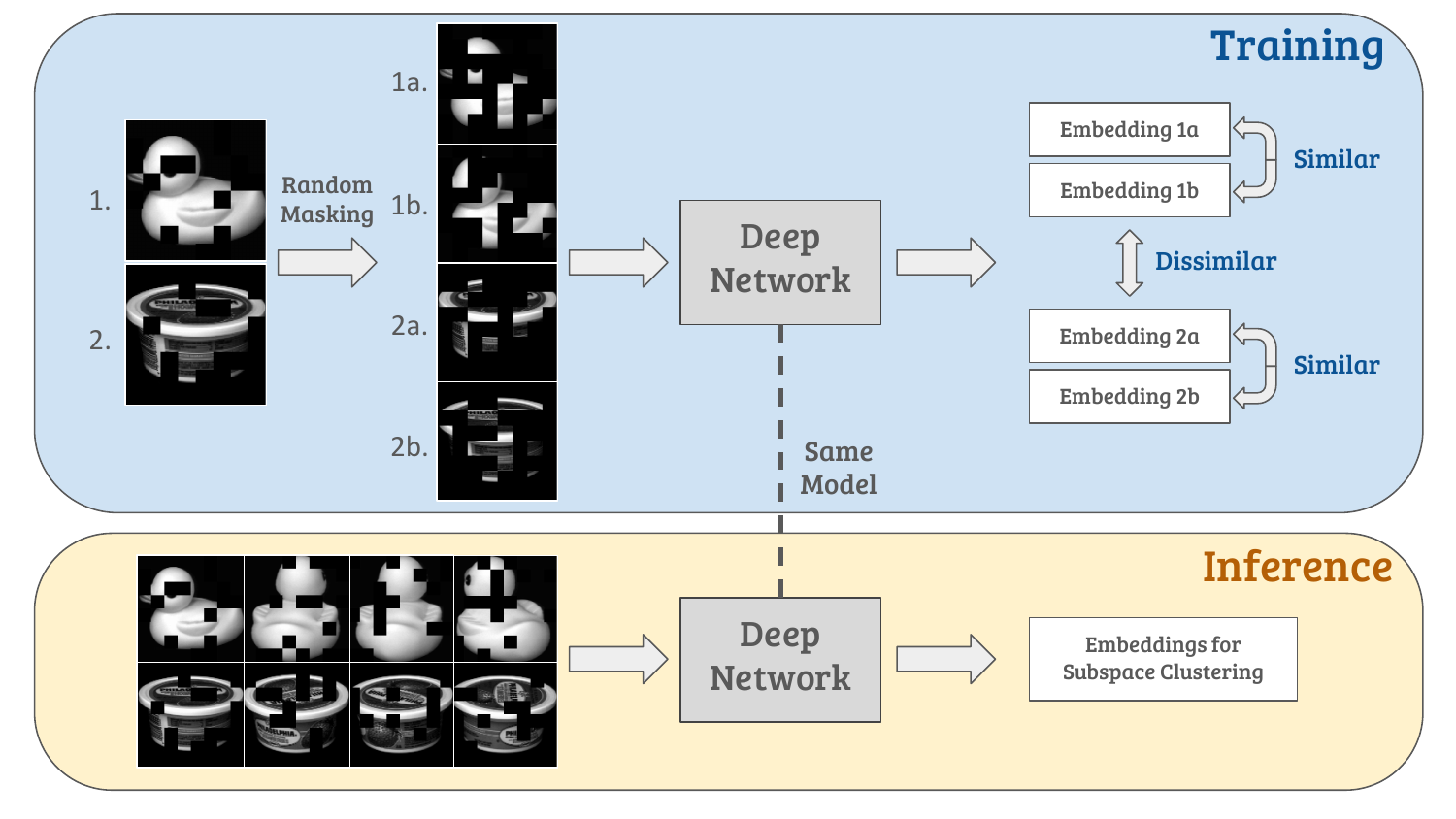}
    \caption{Overview of our Contrastive Subspace Clustering (CSC) pipeline. Two disjoint random masks generate augmented views of each incomplete sample, which are passed through a shared deep network and trained with a SimCLR‐style loss to pull same‐sample embeddings together and push different‐sample embeddings apart. At inference, the original incomplete data is embedded and fed to a subspace‐clustering algorithm.}
    \label{fig:enter-label}
\end{figure*}

In this paper, we propose a novel method for subspace clustering on incomplete data using contrastive self-supervised learning. An overview of our proposed framework is illustrated in Figure~\ref{fig:enter-label}. Our approach leverages random masking to generate augmented views of each partially observed sample. These views are processed by a shared deep neural network, and trained using a SimCLR-style contrastive loss that encourages embeddings of the same original sample to be close, while pushing apart embeddings from different samples. After training, the learned backbone generates embeddings for the original incomplete data, which are then clustered using a sparse subspace clustering algorithm.

\textbf{Organization.}
In Section~\ref{sec:background}, we provide background on recent developments in subspace clustering for both complete and incomplete data, covering traditional methods as well as deep learning-based approaches. Section~\ref{sec:model} introduces and explains our proposed Contrastive Subspace Clustering (CSC) model. In Section~\ref{sec:simulation}, we conduct a comprehensive ablation study on synthetically generated data to examine the effects of model depth, residual connection design, and compare CSC with existing subspace clustering techniques. Finally, Section~\ref{sec:experiment} presents empirical results comparing our model against eight baseline methods across six image datasets under varying sampling rates.

%% file: 2-Background.tex
\subsection{Subspace Clustering}
Subspace clustering is a fundamental problem in unsupervised learning that aims to partition a set of high-dimensional data points into clusters, where each cluster lies near a low-dimensional linear or affine subspace. Unlike traditional clustering methods such as k-means, which assume data clusters around distinct centroids, subspace clustering is designed for data that is better explained by union-of-subspace structures. This formulation is particularly relevant in applications where data naturally lies in multiple low-dimensional manifolds, including gene expression analysis \cite{lan2024jlonmfsc, domeniconi2004subspace, kailing2004density, zheng2020adaptive}, recommendation systems \cite{agarwal2005research, zhang2012guess, cui2021new}, image and video processing \cite{guan2024contrastive, tierney2014subspace, elhamifar2013sparse, peng2020deep}, network security anomaly detection \cite{pu2020hybrid, dromard2016online}, and financial data analytics \cite{pavlidis2006financial, li2021integrated}.

A wide variety of algorithms have been developed to solve this problem. Among the most prominent are methods based on self-expressiveness, where each data point is represented as a sparse or low-rank linear combination of other points in the dataset. Notable examples include Sparse Subspace Clustering (SSC) \cite{elhamifar2013sparse}, which solves a Lasso problem to enforce sparsity, and Low-Rank Representation (LRR) \cite{liu2010robust}, which seeks a globally low-rank coefficient matrix. These methods typically construct an affinity matrix from the learned coefficients and apply spectral clustering to recover the final clusters. More recent developments have enhanced these approaches through the introduction of regularization techniques, such as Elastic Net \cite{you2016oracle} and Elastic Stars \cite{li2024group}, which improve the connectivity of the coefficient matrix and lead to more robust and accurate clustering performance.
Another class of techniques includes algebraic methods, such as Generalized Principal Component Analysis (GPCA) \cite{vidal2005generalized}, and geometric methods, like K-subspaces \cite{agarwal2004k, bradley2000k}, which iteratively alternate between subspace assignment and subspace update. Although effective in certain regimes, these methods are often sensitive to initialization, noise, and the presence of outliers.

\subsection{Deep Learning for Subspace Clustering}
While traditional subspace clustering methods offer strong theoretical guarantees, they often face challenges in real-world settings due to poor scalability, sensitivity to noise, and limited ability to capture nonlinear structures. As datasets grow larger and more complex, classical methods based on pairwise affinities or convex optimization become increasingly inefficient and less effective.

A key advancement in addressing these limitations involves deep autoencoder-based models that jointly learn latent representations and subspace structures. For example, the Deep Subspace Clustering Network (DSC) integrates a self-expressiveness layer within an autoencoder to preserve subspace structure during training \cite{ji2017deep}. Variants like DEC \cite{xie2016unsupervised} and EDESC \cite{cai2022efficient} further improve scalability and robustness by optimizing latent clustering objectives and reducing computational complexity. Similarly, Deep Double Self-Expressive Subspace Clustering (DDSESC) introduces two self-expressiveness objectives—on both encoder and decoder layers—to enhance subspace preservation and global representation consistency \cite{zhao2023deep}.

Beyond autoencoders, more task-specific architectures have been proposed. Contrastive Multi-View Subspace Clustering uses graph convolutional networks to integrate spatial-spectral and textural features in hyperspectral images, achieving state-of-the-art performance via contrastive learning and attention-based fusion \cite{guan2024contrastive}. Adaptive Graph Convolutional Subspace Clustering (AGCSC), have incorporated deep graph structures into subspace affinity learning, resulting in stronger locality preservation and improved clustering accuracy on visual datasets \cite{wei2023adaptive}.
FLNNSC leverage shallow nonlinear transformations to balance linear and nonlinear structure for efficient and accurate clustering across diverse datasets \cite{shi2024nonlinear}. In addition, SENet \cite{zhang2021learning} learns a parametric self-expressiveness function, avoiding large-scale optimization and enabling generalization to unseen data.
More recently, deep subspace models like PRO-DSC have combined self-expressive losses with structured latent representation learning to ensure orthogonal subspace separation and improve robustness \cite{meng2025exploring}. 

Despite these advances, most deep subspace clustering methods still assume fully observed inputs and perform poorly with missing data. Models trained on incomplete samples often overfit to observed entries or rely on biased imputations. Furthermore, many frameworks require pretraining or supervision, limiting their applicability in sparse or noisy domains.

\subsection{Subspace Clustering with Missing Data}
While classical subspace clustering methods perform well under ideal conditions, their effectiveness declines in the presence of missing data—a common issue in real-world applications due to sensor failures, occlusion, irregular sampling, or privacy constraints. Most algorithms assume fully observed inputs to compute affinities or distances, making them unreliable when data is incomplete.
A common workaround is a two-stage process: impute missing entries using methods like matrix completion or k-nearest neighbors, then apply standard clustering. However, this often introduces artifacts that distort the subspace structure, especially at high missing rates or in noisy settings, as the imputation step ignores the underlying geometry.

To address this, joint models have been proposed that combine clustering and missing data handling. EWZF-SSC \cite{yang2015sparse} adapts SSC by zero-filling missing entries and modifying the objective. Methods like GSSC and MSC \cite{pimentel2016group} integrate matrix completion with self-expressiveness, but they rely on large-scale convex optimization and can be sensitive to hyperparameters. More recent approaches, such as Masked Subspace Clustering and SRCC \cite{deng2023semi}, incorporate masking or partial supervision, yet often still assume global low-rank structure or require full affinity matrices—assumptions that break down in high-rank or large-scale settings.

To improve scalability and robustness, recent advances propose new formulations that directly model missingness and adapt to data geometry. Fusion-style subspace clustering \cite{johnson2024fusion} assigns each point to its own subspace and fuses similar ones, offering a natural model selection path and robustness to high-rank data. Similarly, Deep-Union Completion \cite{baskar2025deep} jointly learns imputation and clustering without pretraining, leveraging a union-of-subspaces assumption. Integer Programming for SCMD \cite{soni2024integer} introduces a discrete optimization framework that dynamically selects subspaces and achieves state-of-the-art performance, particularly at high missing rates.
Further, multi-view and deep learning-based approaches have emerged to handle structured or high-dimensional missing data. Robust Multi-View Subspace Clustering \cite{mao2025robust} integrates spatial and spectral cues across views while preserving local structure. RecFormer \cite{liu2023information} uses a two-stage autoencoder with attention to jointly recover and cluster multi-view data. In time-series, SLAC-Time \cite{ghaderi2023self} leverages self-supervised forecasting as a proxy task to learn robust representations without explicit imputation.

Despite these advances, subspace clustering with missing data remains challenging—especially under high missingness or in high-dimensional regimes—where scalability, robustness, and adaptability to structure remain active areas of research.

\subsection{Self-Supervised and Contrastive Learning for Incomplete Data}
Self-supervised learning (SSL) has become a powerful approach for learning meaningful representations from unlabeled data. Among SSL techniques, contrastive learning stands out for its ability to learn discriminative embeddings by encouraging similarity between augmented views of the same instance while distinguishing different ones. Frameworks like SimCLR \cite{chen2020simple}, MoCo \cite{he2020momentum}, and BYOL \cite{grill2020bootstrap} have demonstrated that contrastive learning can match or outperform supervised methods on various downstream tasks.

Contrastive learning is especially well-suited for incomplete data. Unlike reconstruction-based methods that require imputing missing values, contrastive objectives leverage invariance across masked or corrupted views—naturally accommodating partial observations. This label-free training makes it particularly appealing for clustering in unlabeled or semi-structured domains.
Several recent studies explore this intersection. CL-Impute \cite{shi2023cl} uses contrastive learning to recover single-cell RNA-seq data via masked augmentations. CounterCLR \cite{wang2023counterclr} applies causal contrastive modeling to handle non-random missingness in recommendation systems. Though effective, these methods are domain-specific and focus on imputation or prediction rather than clustering.

Beyond contrastive methods, deep learning has been applied to matrix completion. DMF \cite{fan2018matrix} jointly optimizes latent factors and network weights to model complex structures. GMC \cite{monti2017geometric} combines multi-graph convolutions and recurrent networks for structured recommendation data. MHCL \cite{li2024multi} further integrates contrastive learning with hypergraph modeling to capture high-order dependencies. While impactful, these approaches emphasize prediction rather than unsupervised representation learning or clustering.
Other efforts, like CIVis \cite{xie2023towards}, incorporate contrastive learning into visual analytics workflows, enabling user-driven modeling. However, they rely on human input and lack standardized benchmarks for clustering performance.

In contrast, our work is the first to propose a contrastive self-supervised framework for subspace clustering with missing data. By treating random masks as augmentations, our method learns invariant embeddings that capture subspace structure—without reconstruction, imputation, or labels. These embeddings can be directly used with subspace clustering algorithms, offering a scalable and robust solution even under high missing rates.

%% file: 3-Model.tex
In this section, we formally define our novel method of using contrastive learning to solve the subspace clustering on missing data. Our approach leverages data augmentation via random masking and a SimCLR \cite{chen2020simple} contrastive objective to learn embeddings that are well‐suited for downstream subspace clustering.

\textbf{Problem Definition.}
Let $\mathbf{X} \in \mathbb{R}^{d\times n}$ be a data matrix whose columns lie in a union of k low-dimensional subspaces $\{\mathcal{S}_j\}_{j=1}^k$. Each subspace $\mathcal{S}_j$ is of dimension at most $r$, with $r \ll d$. Specifically, the matrix $\mathbf{X}$ can be expressed as:
$\mathbf{X} = [\mathbf{X}_1, \mathbf{X}_2, \dots, \mathbf{X}_k]$,
where each submatrix $\mathbf{X}_j \in \mathbb{R}^{d \times n_j}$, consists of columns exclusively drawn from subspace $\mathcal{S}_j$. 
While each column of $\mathbf{X}$ individually belongs to a low-dimensional subspace, the union of these subspaces $\bigcup_{j=1}^{k}\mathcal{S}_j$ can result in the overall rank of $\mathbf{X}$ being significantly larger, potentially approaching $kr$.
We observe only an incomplete version of the data:
$\mathbf{Y} = \mathbf{M} \odot \mathbf{X},$
where $\mathbf{M} \in \{0,1\}^{d\times n}$ is an unknown binary mask indicating observed entries ($\mathbf{M}_{ij}=1$) and missing entries $(\mathbf{M}_{ij}=0)$, and $\odot$ denotes element-wise multiplication.

The primary goal of the subspace clustering task is to recover the underlying column subspace structure, thereby accurately identifying the partition of columns into their respective low-dimensional subspaces $\{\mathcal{S}_j\}_{j=1}^k$.

\textbf{Augmentation Strategy.}
To enable contrastive learning on incomplete data, for each observed column $\mathbf{y}_i = \mathbf{M}_i \odot \mathbf{x}_i$, we sample two \emph{disjoint} binary masks $\mathbf{M}_i^a$ and $\mathbf{M}_i^b$ such that
$
\mathbf{M}_i^a \odot \mathbf{M}_i^b = \mathbf{0}, 
\quad
\mathbf{M}_i^a + \mathbf{M}_i^b \le \mathbf{M}_i.
$
Applying these masks yields two views
$
\mathbf{y}_i^a = \mathbf{M}_i^a \odot \mathbf{x}_i,
\quad
\mathbf{y}_i^b = \mathbf{M}_i^b \odot \mathbf{x}_i,
$
which serve as positive pairs for contrastive training.

\textbf{Model Architecture.}
We adopt the SimCLR framework \cite{chen2020simple}.  Let $f_\theta:\mathbb{R}^d\to\mathbb{R}^p$ be a deep neural network (the \emph{backbone}) parameterized by $\theta$, and let $g_\phi:\mathbb{R}^p\to\mathbb{R}^q$ be a projection head.  Given an augmented view $\mathbf{y}_i^*$, we compute its representation and projection as
\[
\mathbf{h}_i^* = f_\theta(\mathbf{y}_i^*), 
\quad
\mathbf{z}_i^* = g_\phi\bigl(\mathbf{h}_i^*\bigr).
\]

\textbf{Contrastive Loss.}
We train on a minibatch of $N$ original samples, yielding $2N$ projections $\{\mathbf{z}_i^a,\mathbf{z}_i^b\}_{i=1}^N$.  For each positive pair $(i,a)\!-\!(i,b)$, we define the normalized temperature‐scaled cross‐entropy (NT‐Xent) loss:
\begin{align}
\ell_{i} = &-\log \frac{\exp\!\bigl(\operatorname{sim}(\mathbf{z}_i^a,\mathbf{z}_i^b)/\tau\bigr)}
{\sum_{(j,s)\neq (i,a)}\exp\!\bigl(\operatorname{sim}(\mathbf{z}_i^a,\mathbf{z}_j^s)/\tau\bigr)}
-\log \frac{\exp\!\bigl(\operatorname{sim}(\mathbf{z}_i^b,\mathbf{z}_i^a)/\tau\bigr)}
{\sum_{(j,s)\neq (i,b)}\exp\!\bigl(\operatorname{sim}(\mathbf{z}_i^b,\mathbf{z}_j^s)/\tau\bigr)},
\end{align}
where
$
\operatorname{sim}(\mathbf{u},\mathbf{v}) = \frac{\mathbf{u}^\top\mathbf{v}}{\|\mathbf{u}\|\|\mathbf{v}\|},
$
and $\tau>0$ is a temperature hyperparameter.  The total loss is
$$
\mathcal{L} = \frac{1}{2N}\sum_{i=1}^N \ell_{i}.
$$

\textbf{Inference and Clustering.}
After training, we discard the projection head and use only the trained backbone $f_\theta$ to embed each {original} incomplete sample $\mathbf{y}_i = \mathbf{M}_i\odot\mathbf{x}_i$ into a latent representation
$\widehat{\mathbf{h}}_i = f_\theta(\mathbf{y}_i)$.
The resulting embeddings $\{\widehat{\mathbf{h}}_i\}_{i=1}^n$ capture the subspace structure robustly, even in the presence of missing data. These embeddings can then be directly clustered using any standard clustering algorithm, such as k-means \cite{ahmed2020k}, spectral clustering \cite{von2007tutorial}, or sparse subspace clustering (SSC) \cite{elhamifar2013sparse}, to obtain the final partitioning of data points into their respective low-dimensional subspaces. In our experiments, we choose Sparse Subspace Clustering (SSC) due to its consistent and superior performance in handling linear subspace structures embedded within learned embeddings.

%% file: 4-SyntheticSimulation.tex
In this section, we conduct a comprehensive synthetic study to evaluate our model’s robustness across a variety of architectural and data‐regime configurations. Specifically, we investigate the effects of network depth ($L$), training set size ($N$), and the inclusion or omission of residual connections. We then benchmark our approach against leading missing data subspace clustering algorithms under both low‐ and high‐noise conditions.

\begin{figure}
\setlength{\tabcolsep}{0pt} 
\centering
\begin{tabular}{cc}
  \includegraphics[height=0.22\textwidth, clip, trim={0 0 0 0.7cm}]{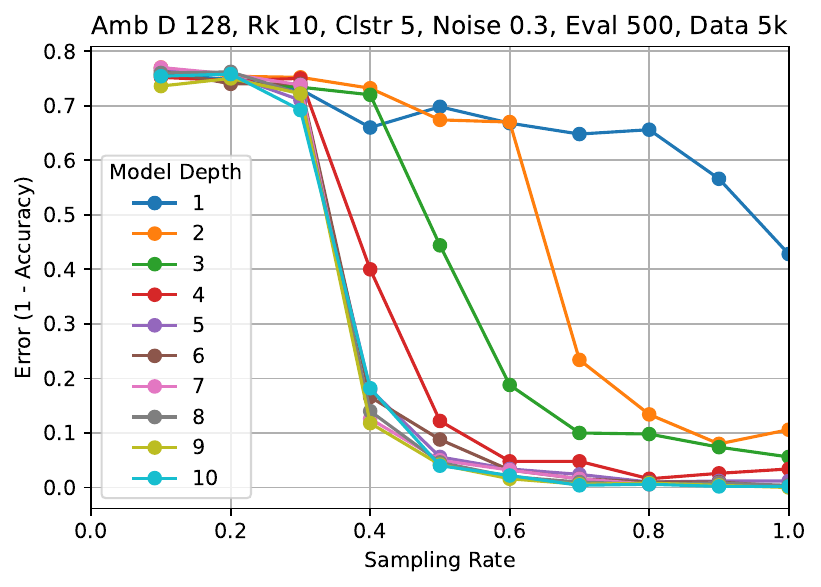} &
  \includegraphics[height=0.22\textwidth, clip, trim={0.7cm 0 0.25cm 0.6cm}]{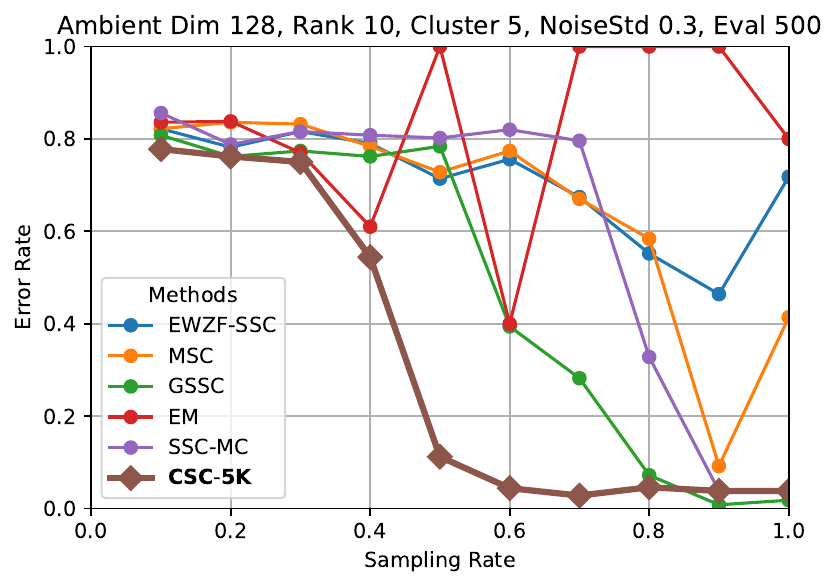} \\
    (a)
  &
    (b)
\end{tabular}
\caption{(a) Clustering error vs.\ sampling rate \(\rho\) for models of depth with \(N = 5000\). (b) Clustering error vs.\ sampling rate \(\rho\) for \(N=5000\) under noise (\(\sigma=0.3\)).}
\label{fig:depth_performance&comparison_classic_hrmc_partial}
\end{figure}

\textbf{Data Generation.}
To synthesize each dataset, we fix the number of subspaces to $k = 5$, each of intrinsic dimension $r = 10$, embedded in an ambient space of dimension $d = 128$.  We draw orthonormal bases 
$
\mathbf{U}_j\in\mathbb{R}^{d\times r},
$
for each subspace $j$. Within subspace $j$, we generate $N/k$ samples via
$
\mathbf{x}_i = \mathbf{U}_{c(i)}\,\mathbf{a}_i,
$
where $\mathbf{a}_i\sim\mathcal{N}(\mathbf{0},\mathbf{I}_r)$ and $c(i)\in\{1,\dots,k\}$ denotes the true subspace index.  To model measurement noise, we add
$
\boldsymbol{\eta}_i\sim\mathcal{N}(\mathbf{0},\sigma^2\mathbf{I}_d),
$
with $\sigma\in\{0.1,\,0.3\}$ controlling the noise level.  Finally, we introduce a \emph{sampling rate} parameter $\rho\in(0,1)$ and generate each binary mask
$
M_{i\ell}\;\sim\;\mathrm{Bernoulli}(\rho),$ for $\ell=1,\dots,d,$
to obtain the observed incomplete sample
$
\mathbf{y}_i = \mathbf{M}_i\odot(\mathbf{x}_i + \boldsymbol{\eta}_i)\in\mathbb{R}^d.
$
Collecting $\{\mathbf{y}_i\}_{i=1}^N$ yields the incomplete data matrix $\mathbf{Y}\in\mathbb{R}^{d\times N}$.  

\begin{figure*}
\setlength{\tabcolsep}{0pt} 
  \centering
  \begin{tabular}{ccc}
    \includegraphics[height=0.22\textwidth, clip, trim = {0 0 0.25cm 0.6cm}]{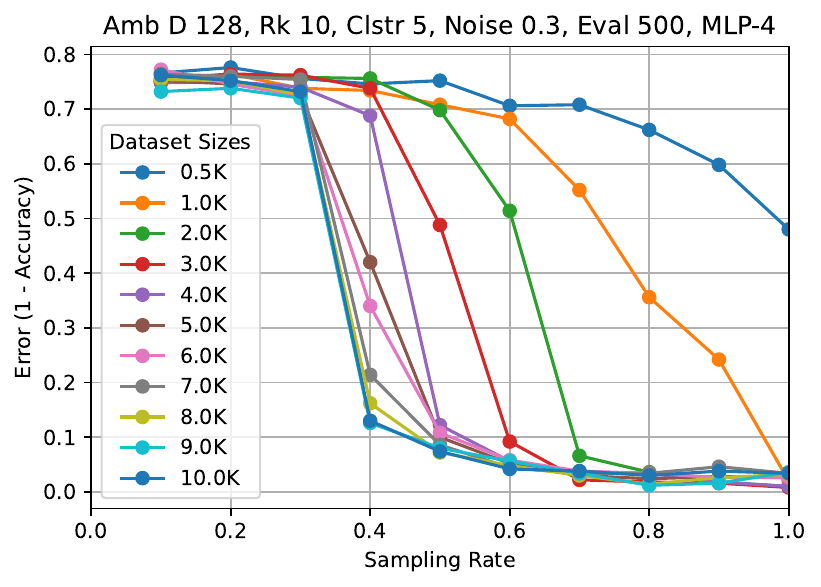} &
    \includegraphics[height=0.22\textwidth, clip, trim = {0 0 1.3cm 0.6cm}]{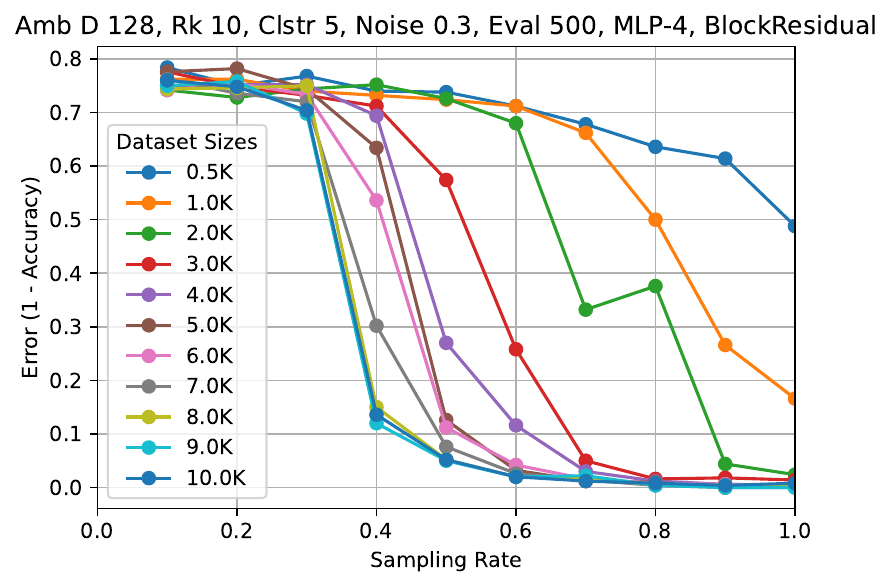} &
    \includegraphics[height=0.22\textwidth, clip, trim = {0 0 0.25cm 0.6cm}]{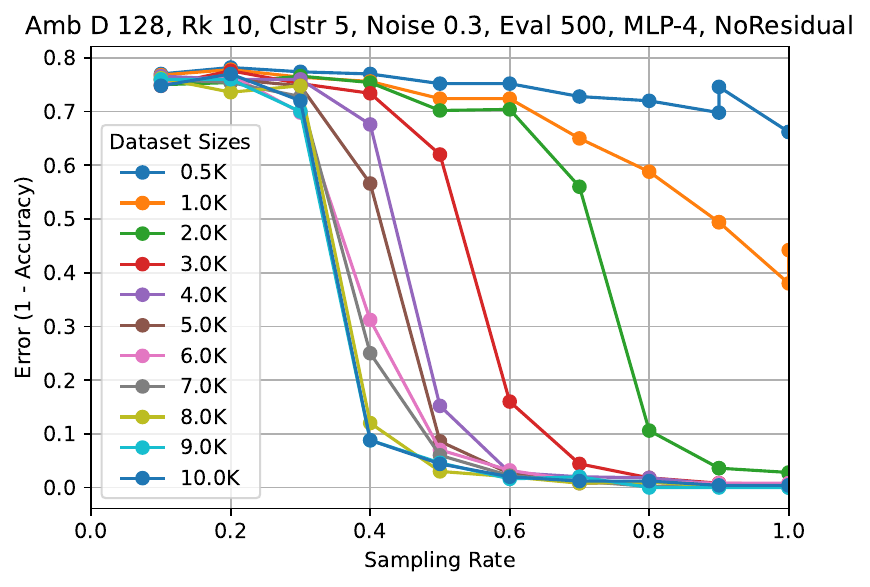} \\
    (a) Full Residual &
    (b) Block‐Level Residual &
    (c) No Residual
  \end{tabular}
  \caption{Clustering error as a function of dataset size \(N\) and sampling rate \(\rho\) for the three residual‐connection configurations: (a) Full residual, (b) Block Residual, and (c) No Residual.}
  \label{fig:residual_comparison}
\end{figure*}

\textbf{Model Configuration.}
Our embedding network consists of $L$ fully‐connected layers of width $d$, each followed by ReLU activations. To facilitate a fair comparison with shallow benchmarking methods, we fix the hidden‐layer dimension equal to the ambient dimension $d$, and we experiment both with and without identity‐mapping residual connections.

\textbf{Evaluation Protocol.}
To assess clustering performance, we uniformly sample 500 incomplete data points from the test set and compute their corresponding latent embeddings using the trained model. These embeddings are then passed to a Sparse Subspace Clustering (SSC) algorithm to produce predicted cluster assignments. Since clustering algorithms may assign arbitrary label indices, we align the predicted clusters with the ground-truth subspace labels by searching over all possible label permutations and selecting the one that yields the highest matching accuracy. The final clustering error is computed as the fraction of misclassified points under this optimal alignment.

\textbf{Model Depth vs.\ Performance.}
In this experiment, we fix the training set size to \(N = 5000\) incomplete samples and evaluate clustering error as a function of the sampling rate \(\rho\) for networks of varying depth \(L\).  Specifically, for each depth \(L\in\{1,2,\dots,8\}\), we train our contrastive embedding model and compute the downstream subspace‐clustering error on held‐out data as \(\rho\) varies in \(\{0.1,0.2,\dots,0.9\}\).
Figure~\ref{fig:depth_performance&comparison_classic_hrmc_partial} (a) illustrates that increasing \(L\) substantially improves clustering accuracy—reflecting greater model expressiveness—but that gains diminish beyond \(L=5\), where performance plateaus.


\textbf{Choice of Residual Connection.}
To encourage the network to learn expressive embeddings while preserving the underlying subspace geometry, we evaluate three residual‐connection strategies: 
1){Full Residual}: identity skip connecting the input of each layer to its output,
2){Block Residual}: skip connections spanning each linear–ReLU block,
3){No Residual}: standard feedforward topology.
For each configuration, we train models of depth $L=5$ on datasets of varying size $N$ and sampling rates $\rho$. We then measure subspace‐clustering error on 500 held‐out samples.
Figure~\ref{fig:residual_comparison} shows that the residual‐connection scheme exerts only a minor influence on performance. At large $N$, the {No Residual} model achieves marginally lower error, whereas at smaller $N$, the {Full Residual} design yields the best results.


\textbf{Comparison with Shallow Methods.}
We fix the number of training samples to \(N = 5000\), and evaluation samples to 500. We compare our CSC approach against standard non‐deep algorithms: EWZF‐SSC \cite{yang2015sparse}, MSC \cite{pimentel2016group}, GSSC \cite{pimentel2016group}, EM \cite{pimentel2014sample}, and SSC‐MC \cite{yang2015sparse}.  Evaluations are conducted under both low‐noise (\(\sigma=0.1\)) and high‐noise (\(\sigma=0.3\)) regimes, with clustering error measured as a function of the sampling rate \(\rho\in\{0.1,0.2,\dots,1.0\}\).

Our findings clearly indicate CSC’s significant advantage over these classical methods, particularly under challenging conditions. Under low noise ($\sigma=0.1$) (Figure~\ref{fig:depth_performance&comparison_classic_hrmc_partial} (b)), CSC consistently achieves lower clustering error across the entire range of sampling rates, reflecting its superior robustness and representational capability. Moreover, the performance gap widens dramatically under high-noise conditions ($\sigma=0.3$)(Figure~\ref{fig:depth_performance&comparison_classic_hrmc_partial} (b)), where classical methods experience rapid performance degradation, especially at lower sampling rates. Conversely, CSC maintains low clustering error even at moderate-to-low sampling rates ($\rho\approx0.5$).

\textbf{Comparison with Masked AutoEncoder.}
We benchmark CSC method against a Masked AutoEncoder (MAE) of comparable capacity (depth \(L=4\)). Both models are trained across a wide range of dataset size ($N$) configurations to assess how effectively each model learns the structure of incomplete samples.  We evaluate clustering error as a function of sampling rate \(\rho\) under low‐noise (\(\sigma=0.1\)) and high‐noise (\(\sigma=0.3\)) regimes.

\begin{figure*}
\setlength{\tabcolsep}{0pt} 
  \centering
  \begin{tabular}{cccc}
    \includegraphics[height=0.176\linewidth, clip, trim = {0 0 0.25cm 0.6cm}]{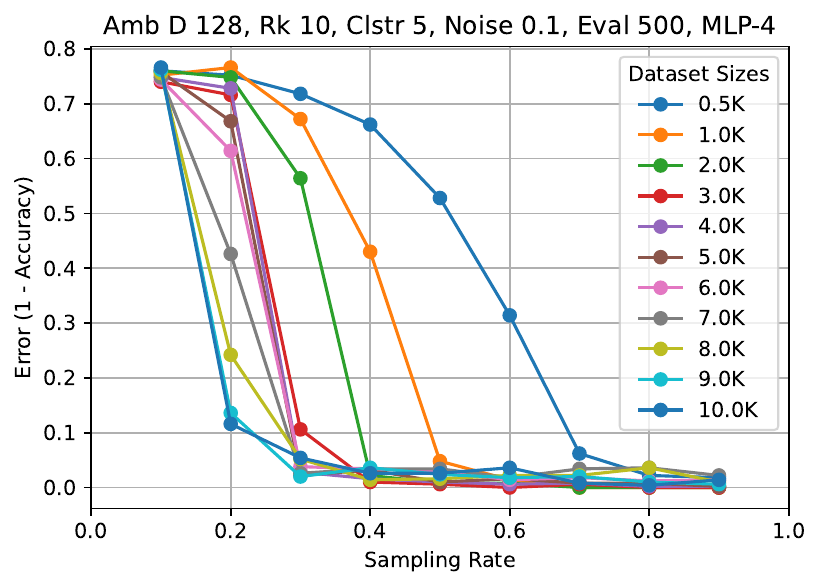} &
    \includegraphics[height=0.176\linewidth, clip, trim = {0.6cm 0 0.25cm 0.6cm}]{Figures/CSC_HighNoise.pdf} &
    \includegraphics[height=0.176\linewidth, clip, trim = {0.6cm 0 0.25cm 0.6cm}]{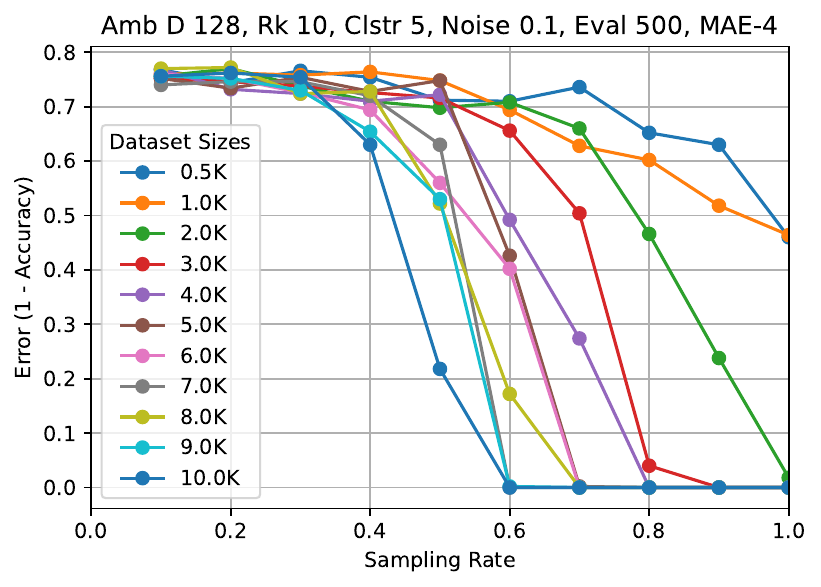} &
    \includegraphics[height=0.176\linewidth, clip, trim = {0.6cm 0 0.25cm 0.6cm}]{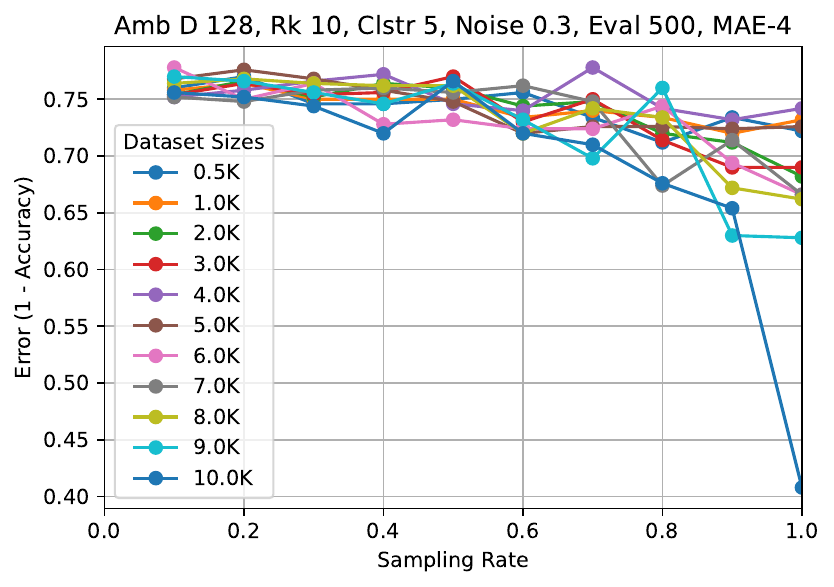} \\[-1ex]
    (a) CSC (Low Noise) & (b) CSC (High Noise) & 
    (c) MAE (Low Noise) &(d) MAE (High Noise)
  \end{tabular}
  \caption{Clustering error versus sampling rate \(\rho\) for CSC and MAE under (a,b) low‐noise \(\sigma=0.1\) and (c,d) high‐noise \(\sigma=0.3\).  CSC consistently outperforms MAE across all \(\rho\) and noise levels.}
  \label{fig:comparison_mae}
\end{figure*}

Results of this comparison are depicted in Figure~\ref{fig:comparison_mae}. Clearly, CSC outperforms MAE consistently across all evaluated sampling rates and under both noise conditions. Under the low-noise scenario (Figures~\ref{fig:comparison_mae}(a) and (c)), CSC achieves lower clustering errors, especially noticeable at lower sampling rates ($\rho < 0.5$). 
Under the high-noise regime (Figures~\ref{fig:comparison_mae}(b) and (d)), the superiority of CSC becomes even more pronounced. While MAE’s performance deteriorates sharply in response to increased noise, CSC maintains stable and markedly lower clustering errors throughout the entire sampling rate range. 

%% file: 5-RealExperiment.tex
In this section, we conduct extensive experiments to evaluate the performance of our proposed Contrastive Subspace Clustering (CSC) framework under realistic settings. Our evaluation includes standard computer vision benchmarks and challenging hyperspectral imaging (HSI) datasets characterized by high dimensionality and structured missingness. We compare CSC against a broad range of state-of-the-art methods, including both classical matrix completion techniques and recent deep learning-based baselines. In what follows, we detail the datasets used, competing methods, model configurations, and evaluation protocol, and present both quantitative results and ablation studies on the batch size.

\textbf{Dataset.}
To validate the effectiveness of our contrastive subspace clustering framework on real-world data, we evaluate two classes of datasets. First, we employ two canonical computer-vision benchmarks: {COIL20~\cite{nene1996columbia}} and {MNIST~\cite{lecun2010mnist}}, both widely utilized for clustering and representation learning evaluation. Second, we test on hyperspectral sensor image (HSI) datasets: {IndianPines}, {Pavia University}, and {Salinas}. These HSI datasets, acquired from geographically diverse regions in the United States and Italy, contain spectral reflectance across hundreds of spectral bands, covering varied scenes such as agricultural fields, urban environments, and mixed vegetation, thus providing challenging real-world conditions for clustering due to spectral similarities and high dimensionality.

\textbf{Method.}
CSC is compared against a broad spectrum of state‐of‐the‐art methods and deep‐learning baselines, including DMFMC~\cite{fan2018matrix}, GSSC~\cite{pimentel2016group}, EWZF‐SSC~\cite{yang2015sparse}, SSC‐MC~\cite{yang2015sparse}, ZF‐SSC~\cite{yang2015sparse}, FLNNSC~\cite{shi2024nonlinear}, Masked AutoEncoder(MAE), and DSC~\cite{ji2017deep}. For each benchmark, all hyperparameters are tuned via cross‐validation on held‐out subsets. The tuned hyperparameters for each method are reported in Appendix. Each algorithm is executed five times at each sampling rate \(\rho\), yielding the average clustering accuracy \(\mathrm{Acc}(\rho)\). We further summarize overall effectiveness by computing the mean accuracy across the set of sampling rates.

\textbf{Backbone Models.}
Deep learning baselines are instantiated using backbone architectures tailored to their respective data modalities: convolutional neural networks for image datasets and one-dimensional convolutional networks for hyperspectral image (HSI) datasets. Additionally, for both modalities, we also report the performance of a simple feedforward neural network (denoted as "MLP10") for further comparison.

For image datasets, our CSC model employs a ResNet-18 architecture as its backbone. In contrast, the Masked AutoEncoder (MAE) and Deep Subspace Clustering (DSC) methods utilize an 18-layer convolutional autoencoder (denote as "Conv18"). Specifically, their encoder comprises convolutional layers with channel dimensions [64, 64, 128, 128, 256, 256, 512, 512, 256], and the decoder mirrors this arrangement in reverse. Each convolutional layer has kernel size 3, stride 1, and padding 1, followed by batch normalization and ReLU activation. Additionally, every two convolutional blocks are succeeded by a max-pooling operation. The final output layer replaces the max pooling with a sigmoid activation function to ensure normalized outputs.

For HSI datasets, all deep learning models employ 10-layer one-dimensional convolutional architectures (denote as "Conv10") to maintain fair comparisons with classical methods that do not leverage spatial or geographical context. Specifically, our CSC method uses a convolutional model with progressively increasing channel dimensions [32, 64, 128, 256, 512, 1024]. Each convolutional layer is followed by one-dimensional batch normalization, ReLU activation, and max-pooling layers. After the convolutional layers, four fully-connected layers with dropout regularization further process the features. Similarly, MAE and DSC methods adopt a comparable 10-layer encoder-decoder architecture with encoder channel dimensions [32, 64, 128, 256, 512] and decoder channel dimensions in reverse order. Each convolutional layer is likewise followed by one-dimensional batch normalization and ReLU activation.

\input{SmallTable-MNIST}

\input{SmallTable-Pavia}

\textbf{Evaluation.}
Clustering accuracy is obtained by randomly sampling 500 examples from each dataset and applying a remapping function to maximize alignment between cluster assignments and ground-truth labels. All methods, except ours and MAE, are restricted to using only these 500 examples for clustering, as they do not involve a training phase. In contrast, CSC and MAE are trained on the full training dataset, while evaluation is performed on the sampled 500 test examples. For DSC, the loss computation is masked to include only the observed entries in the reconstruction loss. Since DSC operates in a single-batch training setting, its batch size is fixed at 500.

\textbf{Result.}
Clustering accuracies are reported in Table~\ref{tab:mnist_acc} for MNIST, Table~\ref{tab:indianpines} for IndianPines, and Table~\ref{tab:pavia} for Pavia. Additional results on full missing rate spectrum with additional datasets can be found in technical Appendix. On MNIST, Pavia, and IndianPines, CSC outperforms all benchmarks by more than 15\%, 5\%, and 6\%, respectively.
For COIL20, PaviaU, and Salinas, CSC performs within 3\% of the state-of-the-art. All CSC experiments are conducted using a single NVIDIA A100 GPU, with training completing in under 3 hours for MNIST and under 30 minutes for each of the remaining datasets.

\input{SmallTable-IndianPines}

\textbf{Ablation on Batch Size.}
We evaluate the performance of CSC under a 50\% missing rate across different batch sizes on MNIST and Pavia, using the best-performing backbone architecture. Results are reported in Appendix. We observe that CSC consistently performs well when the batch size is within the range of 64 to 256.

%% file: SmallTable-MNIST.tex
\begin{table}
\centering
\setlength{\tabcolsep}{5pt}
\begin{tabular}{lrrrrrr}
\toprule
Sampling Rate & 0.1 & 0.3 & 0.5 & 0.7 & 0.9 & Mean \\
Method &  &  &  &  &  &  \\
\midrule
CSC-MLP10 & \textbf{46.40} & \textbf{58.72} & \textbf{72.16} & \textbf{71.92} & \textbf{72.16} & \textbf{65.95} \\
MAE-MLP10 & 21.32 & 57.96 & 52.88 & 54.76 & 59.32 & 50.98 \\
CSC-ResNet18 & 42.00 & 50.56 & 46.44 & 42.64 & 43.16 & 46.18 \\
MAE-Conv18 & 36.64 & 38.40 & 47.76 & 51.44 & 45.04 & 43.65 \\
EWZF-SSC & 19.44 & 29.36 & 40.56 & 48.92 & 49.40 & 39.22 \\
MSC & 19.32 & 28.44 & 39.92 & 48.08 & 49.28 & 38.63 \\
DMFMC & 16.24 & 37.36 & 42.84 & 39.00 & 38.88 & 37.00 \\
GSSC & 19.44 & 27.36 & 35.88 & 44.44 & 44.32 & 35.81 \\
FLNNSC & 17.84 & 27.84 & 30.84 & 36.84 & 35.68 & 30.36 \\
SSC-MC & 16.72 & 17.72 & 27.96 & 36.40 & 39.28 & 28.96 \\
DSC-Conv18 & 19.32 & 19.96 & 23.16 & 25.48 & 22.68 & 23.00 \\
DSC-MLP10 & 17.40 & 18.04 & 21.08 & 21.00 & 23.88 & 21.14 \\
\bottomrule
\end{tabular}
\caption{Clustering Accuracy on MNIST. CSC is our method for Contrastive Subspace Clustering.}
\label{tab:mnist_acc}
\end{table}

%% file: SmallTable-Pavia.tex
\begin{table}
\centering
\setlength{\tabcolsep}{5pt}
\begin{tabular}{lrrrrrr}
\toprule
Sampling Rate & 0.1 & 0.3 & 0.5 & 0.7 & 0.9 & Mean \\
Method &  &  &  &  &  &  \\
\midrule
CSC-Conv10 & 68.28 & 76.16 & \textbf{78.12} & 76.72 & \textbf{77.24} & \textbf{74.28} \\
ZF-SSC & 57.72 & \textbf{78.92} & 76.96 & \textbf{77.56} & 72.52 & 68.88 \\
DSC-Conv10 & \textbf{74.92} & 63.64 & 63.04 & 66.06 & 66.10 & 65.70 \\
DMFMC & 70.12 & 51.36 & 65.88 & 63.64 & 55.12 & 62.11 \\
MAE-Conv10 & 51.68 & 51.20 & 52.12 & 52.36 & 51.40 & 51.74 \\
MAE-MLP10 & 52.48 & 52.36 & 52.04 & 51.12 & 51.20 & 51.48 \\
EWZF-SSC & 28.80 & 40.76 & 46.40 & 56.24 & 67.44 & 47.26 \\
FLNNSC & 31.32 & 26.76 & 15.72 & 27.80 & 76.32 & 39.82 \\
DSC-MLP10 & 40.64 & 23.76 & 31.00 & 42.16 & 60.76 & 39.40 \\
GSSC & 18.44 & 25.92 & 29.04 & 33.80 & 36.35 & 30.01 \\
CSC-MLP10 & 29.88 & 28.08 & 26.48 & 27.96 & 29.68 & 28.50 \\
MSC & 21.28 & 27.32 & 25.56 & 26.96 & 21.12 & 25.38 \\
SSC-MC & 17.08 & 16.64 & 20.00 & 22.20 & 24.32 & 20.67 \\
\bottomrule
\end{tabular}
\caption{Clustering Accuracy on HSI-Pavia with 10 classes. CSC is our method for Contrastive Subspace Clustering.}
\label{tab:pavia}
\end{table}

%% file: SmallTable-IndianPines.tex
\begin{table}
\centering
\setlength{\tabcolsep}{5pt}
\begin{tabular}{lrrrrrr}
\toprule
Sampling Rate & 0.1 & 0.3 & 0.5 & 0.7 & 0.9 & Mean \\
Method &  &  &  &  &  &  \\
\midrule
CSC-Conv10 & \textbf{43.08} & \textbf{49.96} & \textbf{43.04} & \textbf{50.60} & \textbf{44.68} & \textbf{47.29} \\
DMFMC & 41.64 & 43.08 & 40.64 & 41.16 & 39.48 & 40.70 \\
DSC-Conv10 & 38.80 & 41.84 & 37.52 & 41.60 & 33.68 & 38.61 \\
MAE-Conv10 & 35.16 & 36.16 & 37.80 & 36.60 & 36.20 & 36.21 \\
MAE-MLP10 & 36.80 & 35.84 & 36.60 & 36.44 & 35.88 & 36.02 \\
DSC-MLP10 & 35.04 & 22.56 & 25.40 & 28.88 & 28.80 & 27.76 \\
GSSC & 12.72 & 18.32 & 21.76 & 23.60 & 24.70 & 21.77 \\
MSC & 12.72 & 19.68 & 20.12 & 22.04 & 23.08 & 20.75 \\
EWZF-SSC & 16.56 & 15.68 & 16.32 & 21.32 & 25.32 & 19.58 \\
FLNNSC & 12.80 & 12.76 & 12.24 & 12.36 & 27.76 & 18.30 \\
CSC-MLP10 & 12.92 & 14.16 & 17.40 & 19.28 & 20.08 & 17.57 \\
SSC-MC & 12.36 & 13.08 & 13.96 & 20.76 & 23.88 & 17.49 \\
ZF-SSC & 20.56 & 13.28 & 13.60 & 12.72 & 20.08 & 15.92 \\
\bottomrule
\end{tabular}
\caption{Clustering Accuracy on HSI-IndianPines with 17 classes. CSC is our method for Contrastive Subspace Clustering.}
\label{tab:indianpines}
\end{table}